\documentclass[10pt,twocolumn,letterpaper]{article}

\usepackage{3dv}
\usepackage{times}
\usepackage{epsfig}
\usepackage{graphicx}
\usepackage{amsmath}
\usepackage{amssymb}
\usepackage{placeins}
\usepackage{afterpage}
\usepackage{subcaption}

\usepackage[pagebackref=true,breaklinks=true,letterpaper=true,colorlinks,bookmarks=false]{hyperref}

\threedvfinalcopy 

\def\threedvPaperID{****} 

\ifthreedvfinal\fi
\begin{document}

\title{TRANSPR: Transparency Ray-Accumulating Neural 3D Scene Point Renderer}

\author{Maria Kolos$^{1}$\footnotemark \qquad Artem Sevastopolsky$^{1,2}$\footnotemark[1] \qquad Victor Lempitsky$^{1,2}$\\
$^1$ Samsung AI Center, Moscow, Russia\\
$^2$ Skolkovo Institute of Science and Technology (Skoltech), Moscow, Russia\\
\texttt{mariakolos1@gmail.com, \{a.sevastopol, v.lempitsky\}@samsung.com}
}

\twocolumn[{%
\renewcommand\twocolumn[1][]{#1}
    \maketitle
    \begin{center}
            \centering
            \setlength\tabcolsep{0pt}
            \begin{tabular}{ccccc}
                \includegraphics[width=.2\textwidth]{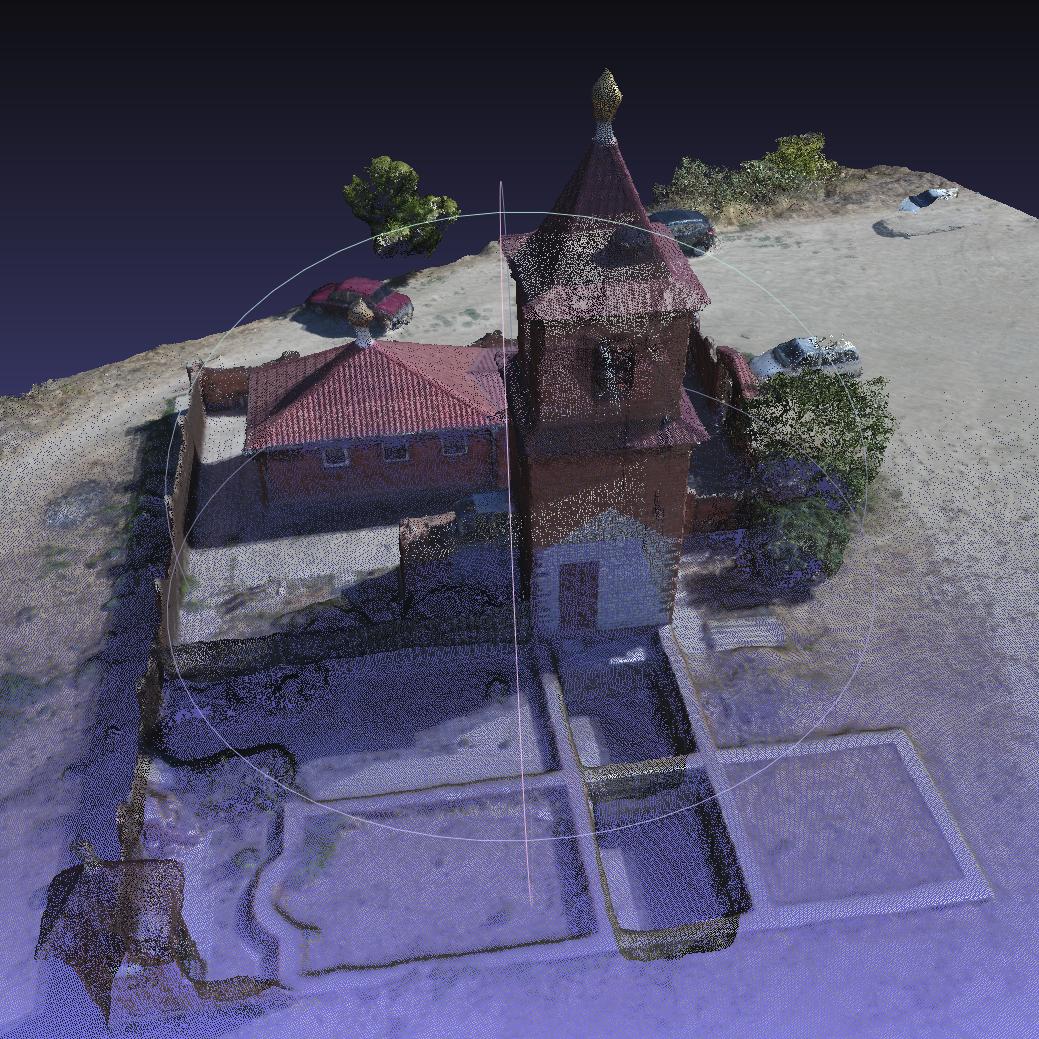} & \includegraphics[width=.2\textwidth]{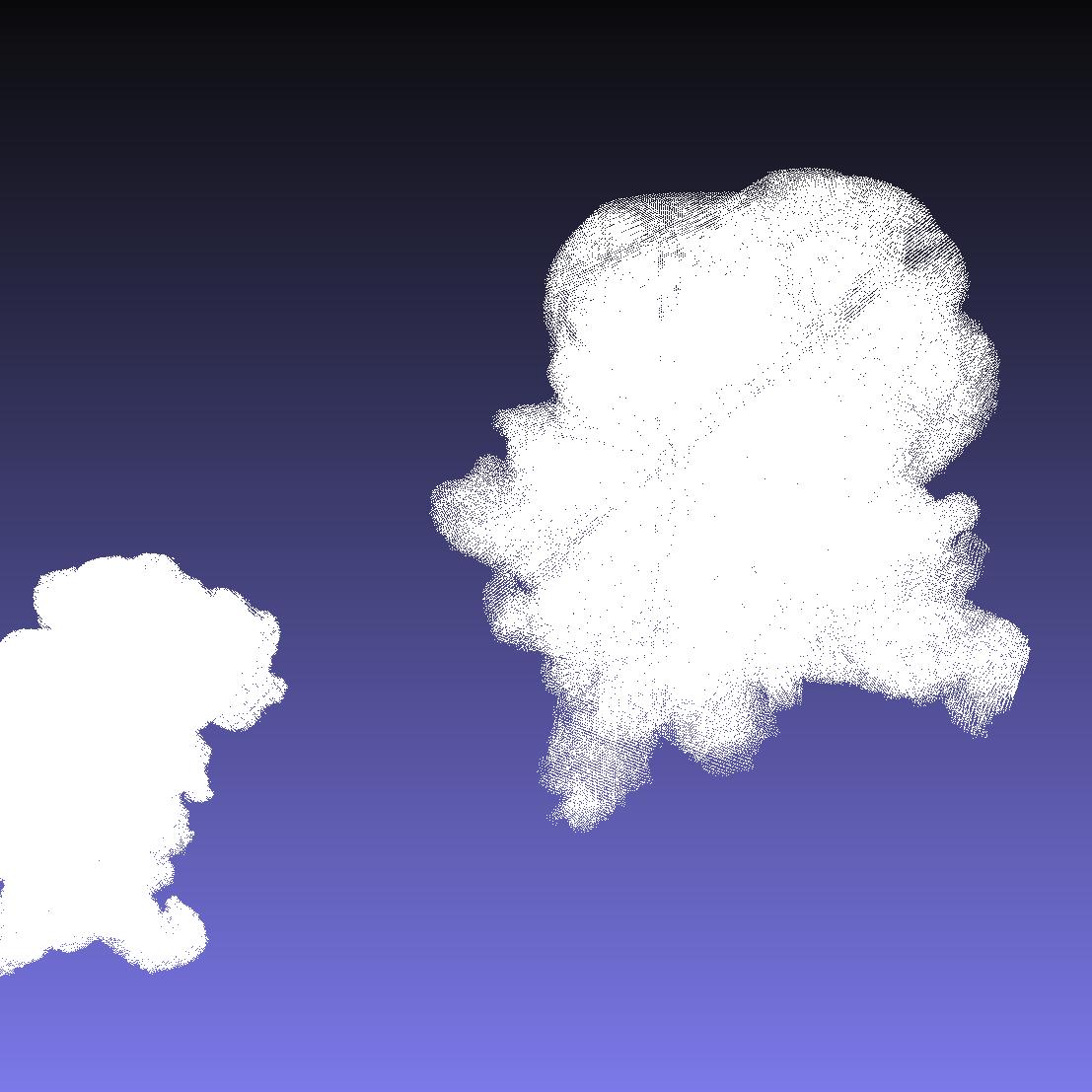} & \includegraphics[width=.2\textwidth]{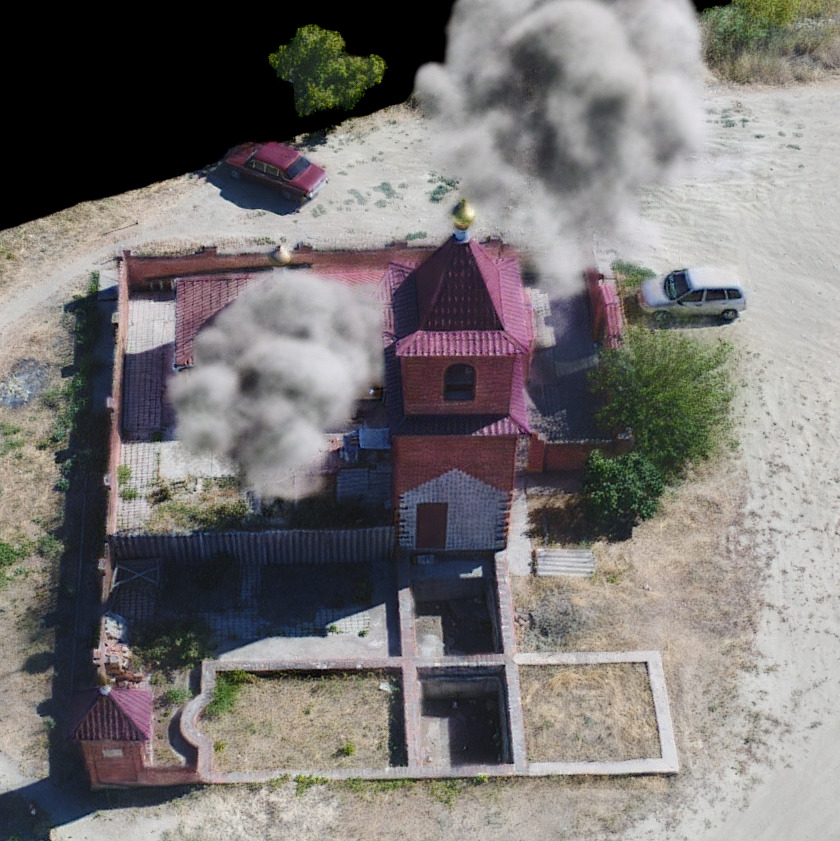} & \includegraphics[width=.2\textwidth]{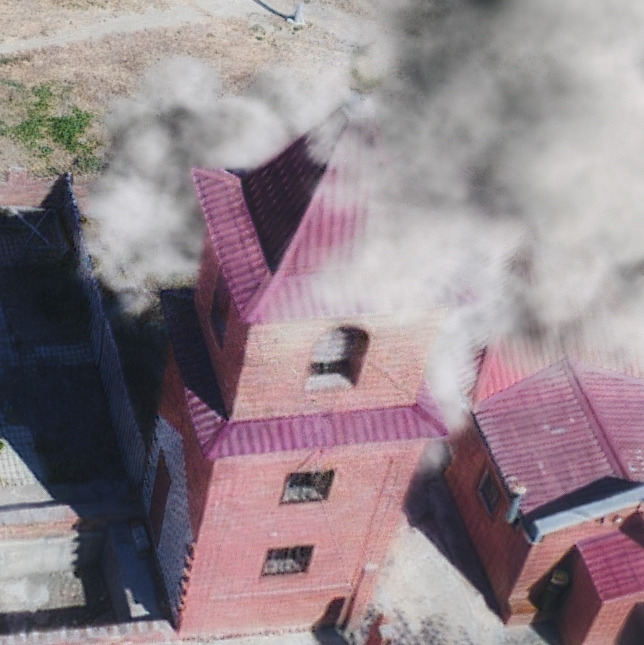} & \includegraphics[width=.2\textwidth]{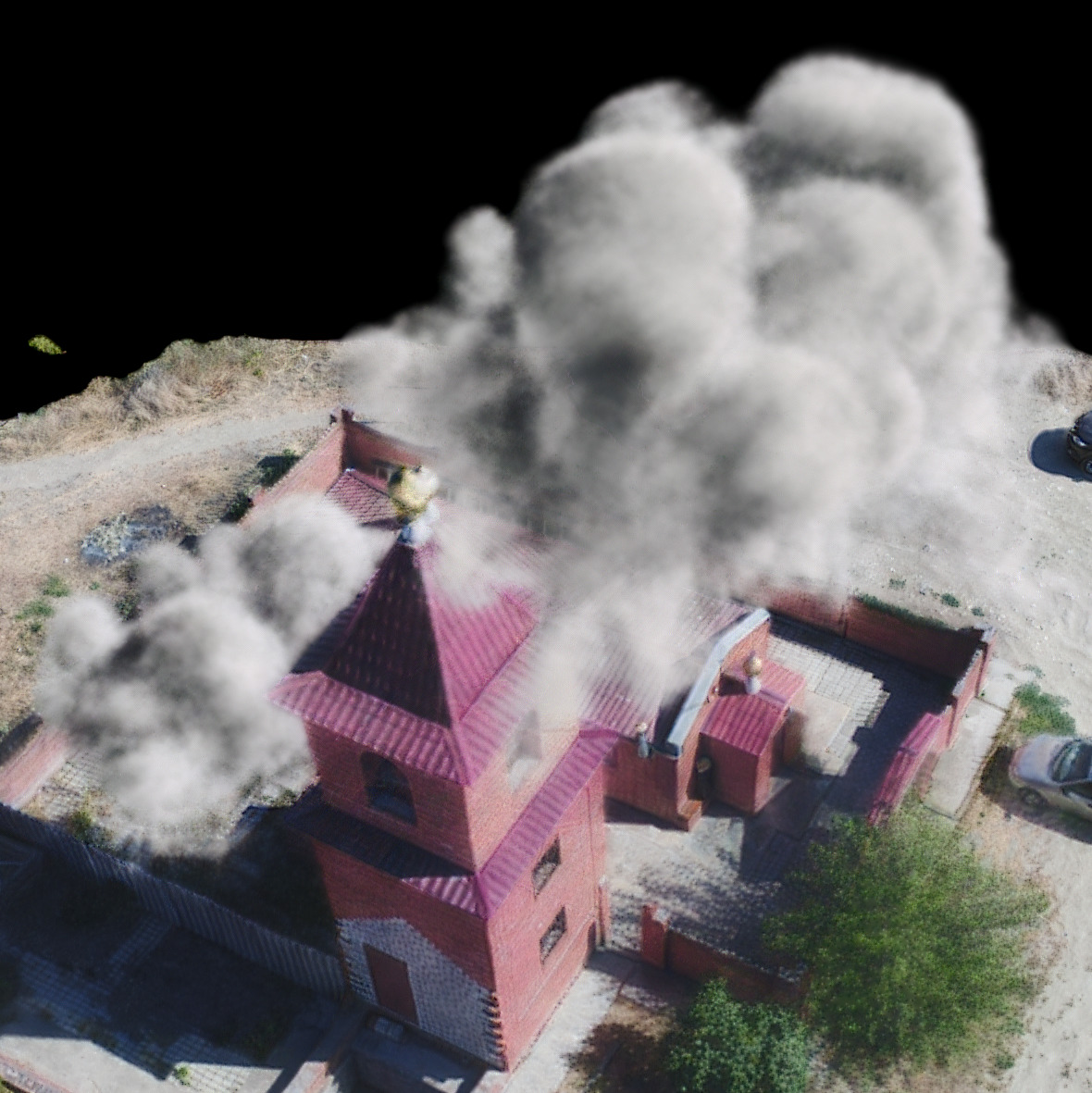} \\
            \end{tabular}
            \captionof{figure}{Given a scene containing semi-transparent objects and represented as a point cloud, our neural pipeline can synthesize photorealistic renderings of this scene from arbitrary, previously unseen viewpoints. The scene can be a comprised of a composition of independent objects. Left-to-right: (1) point cloud for the \textit{Church} scene, (2) point cloud for the \textit{Clouds} scene, (3--5) combined renderings by TRANSPR from 3 viewpoints.}
            \label{fig:teaser}
    \end{center}
}]

\footnotetext[1]{indicates equal contribution}
\begin{abstract}

We propose and evaluate a neural point-based graphics method that can model semi-transparent scene parts. Similarly to its predecessor pipeline, ours uses point clouds to model proxy geometry, and augments each point with a neural descriptor. Additionally, a learnable transparency value is introduced in our approach for each point. 

Our neural rendering procedure consists of two steps. Firstly, the point cloud is rasterized using ray grouping into a multi-channel image. This is followed by the neural rendering step that ``translates'' the rasterized image into an RGB output using a learnable convolutional network. New scenes can be modeled using gradient-based optimization of neural descriptors and of the rendering network.




We show that novel views of semi-transparent point cloud scenes can be generated after training with our approach. Our experiments demonstrate the benefit of introducing semi-transparency into the neural point-based modeling for a range of scenes with semi-transparent parts. 

\end{abstract}
\section{Introduction}

Many recent works investigate how classical computer graphics approaches and representations can be integrated with \textit{neural rendering}~\cite{Tewari20}. Among these approaches, several investigate the use of point clouds for modeling scene geometry~\cite{Bui18,Yifan19,Meshry19,Pittaluga19,Aliev20}. Neural point-based graphics offer an attractive combination of relatively fast and realistic rendering, ability to model scenes with thin parts, robustness towards imperfect camera registration and geometric models.

Previous neural point-based graphics approaches use ``hard'' rasterization of point clouds using the z-buffer algorithm, and are not designed to model scenes with semi-transparent parts. Here, we address this limitation and show how this important class of scenes can be handled by introducing transparency parameters at the level of individual points.

In more detail, we start with the approach of Aliev~et~al.~\cite{Aliev20} that assigns each point a neural descriptor that characterizes local photometric and geometric properties. We then augment point descriptors with alpha values that model points transparency. During rendering, point clouds are rasterized at multiple resolutions using alpha values and front-to-back blending over a ray. Similarly to \cite{Aliev20}, the neural descriptors are used as pseudo-colors during such rasterization. The obtained multichannel image is translated into a realistic image using convolutional rendering network.

Overall, our approach, which we call Transparency Ray-Accumulating Neural 3D Scene Point Renderer (TRANSPR) extends \cite{Aliev20} by augmenting point desriptors with alpha values, and replacing z-buffer rasterization with ray marching (ray accumulation). This augmentation preserves the differentiability of rendering, so that new scenes can be modeled via gradient-based optimization of point descriptors and the rendering network parameters. 

In the experimental comparison, we show that such extension of the neural point-based graphics framework makes it more suitable for scenes with semi-transparent parts. The experiments performed on a number of synthetic and real-world scenes also compare our method to a state-of-the-art neural volumetric rendering approach~\cite{Mildenhall20}.

\section{Related work}

Recently, there is a growing amount of works that combine geometric representations with neural rendering~\cite{Tewari20}. Most representations that have been developed by the computer graphics community have been tried within such representations. Thus, the classical mesh-based geometry representation gave rise to the deferred neural rendering (DNR) approach~\cite{Thies19} that has produced excellent results. 

The most related to ours are works that use point clouds as geometric representations such as neural splatting~\cite{Bui18}, differential surface splatting~\cite{Yifan19}, neural rerendering in the wild~\cite{Meshry19}, and neural point-based graphics (NPBG)~\cite{Aliev20}. These approaches all build on top of the classical point-based graphics works~\cite{Gross02,Kobbelt04} and augment them with neural rendering. The NPBG approach as well as the mesh-based DNR approach both use learnable surface descriptors that are trained together with the rendering network via the backpropagation process. 

The approaches mentioned above all use explicit surface representations (meshes or point clouds), do not model transparency, and as such are not well-suited for modeling scenes with semi-transparent parts. In contrast, a growing group of methods is based on volumetric modeling combined with neural rendering. Thus, the Deep Voxels~\cite{Sitzmann19a} and the Neural Volumes approaches~\cite{Lombardi19} use regular rectangular voxel grids to model the scenes, while more recently methods that use implicit scene modeling using neural networks~\cite{Sitzmann19b,Mildenhall20} have been proposed. All these approaches are well-suited to model complex scenes with semi-transparency, however they suffer from slow rendering time due to the need to integrate the neural descriptors along rays. 

Our approach (TRANSPR) sits in-between the point-based and the volumetric neural rendering approaches, as it relies on a point cloud to model scene geometry, while using ray integration (accumulation) to obtain neural descriptors that are submitted into the rendering network. Experimentally, we show that it attains a good quality-speed tradeoff, being able to model the scenes with semi-transparency better than the NPBG system it is based upon, while being faster than volumetric neural rendering approaches such as NeRF~\cite{Mildenhall20}.

Finally, our work is related to a class of scene modeling approaches based on multiplane images~\cite{Szeliski98,Zhou18,Flynn19} that similarly use back-to-front accumulation of semi-transparent elements. Unlike those works that allow limited viewpoint variation for outside-in scene modeling, we focus on the modeling of closed scenes from a wider variation of viewpoints.

\section{Method}
\def\P{\mathbf{P}}
\def\D{\mathbf{D}}
\def\C{C}
\def\W{W}
\def\H{H}
\def\Loss{\mathcal{L}}
\def\s{\mathbf{s}}
\def\Rend{\mathcal{R}}
\def\I{\mathbf{I}}

\subsection{Neural Point-Based Graphics}
Our method extends the Neural Point-Based Graphics (NPBG)~\cite{Aliev20} rendering framework, which aims to realistically model the appearance of a scene represented as a point cloud. For the sake of completeness, we first briefly outline how the NPBG method works.



Let $\P = \{p_1,p_2, \dots,p_N\}$, where $p_j = \{x_j, y_j, z_j\}$, be a point cloud comprising the scene to be rendered. It is expected that points of $\P$ are allocated densely enough on the surfaces of all objects in the scene; most often, such representation is given by a Structure-from-Motion (SfM) or 3D reconstruction method of any kind. Let us attach an $M$-dimensional descriptor to each point $d_i$ ($\D = \{d_1,d_2,\dots,d_N\}$) that is initialized randomly and will later be learned. 

Given a pinhole camera $\C$ parameterized by predefined extrinsic and intrinsic parameters, the rendering process starts with \textit{rasterization} of descriptors $\D$ onto a set of $M$-channel \textit{raw images} $S[0], S[1], \dots, S[T]$, where $S[t] = S[t](\P, \D, C)$ is of spatial size $\frac W{2^t} \times \frac H{2^t}$. For each $S[t]$, this step is implemented via hard z-buffering: for each point, the screen-space coordinates $x, y$ are first calculated via perspective projection, and afterwards, for each integer output image pixel $(m, n)$, the point $p_i \in \P$ closest to the world location of the camera $C$ is selected among all those points which project to $(m, n)$ onto a chosen image $S[t]$. The descriptor of this point is put into the respective pixel of a \textit{raw image} $S[t](\P, \D, C)[m, n] = d_i$. Less formally, during rasterization, renderings in several resolutions of the point cloud $\P$, where all points are colored with their descriptors, are produced from the camera viewpoint. Since the point cloud does not contain any topology information, the reason behind using a pyramid of images in NPBG instead of one image is to fight \textit{surface bleeding}. Higher resolution maps have more details but have holes and are thus prone to bleeding. Lower resolution raw images have fewer holes but lack detailization (see~\cite{Aliev20} for further discussion).


Multiscale raw images are passed as inputs to the rendering network $\Rend_{\theta}$, the architecture of which is based on a wide-spread U-Net~\cite{Ronneberger15} with several modifications made. Most importantly, $\Rend_\theta$ stacks each raw image with the first feature map of the respective level of resolution. A final RGB image $I_{W \times H}$ is returned by $\Rend_\theta$:

\begin{align}
    I_{\W \times \H} = \Rend_{\theta}(S[0], \dots  S[T]),
    \label{eq:rendering_net}
\end{align}

The rasterization step is implemented via OpenGL, and the pipeline allows for the gradient flow from the loss function to the projected descriptors. $T=4$ was taken in all the experiments. 

The model is fitted to data via  backpropagation w.r.t.\ learnable parameters $\theta$ of the rendering network and the point descriptors $\D$. The system is trained on $K$ scenes, each represented by a point cloud $\P^k$ (equipped with the descriptors $\D^k$ to be learned) and the set of $L_k$ training ground truth RGB images $\I^k = \{I^{k,1},I^{k,2},\dots I^{k,L_k}\}$, as well as the respective camera parameters $\{\C^{k,1},\C^{k,2},\dots \C^{k,L_k}\}$. The objective $\Loss$ is expressed by the mismatch $\Delta$ between the rendered and the ground truth RGB images:

\begin{equation}
    \begin{aligned}
    & \Loss(\theta,\D^1,\D^2,\dots,\D^K) =  \\
                                       & \sum_{k=1}^K\sum_{l=1}^{L_k} \Delta\left(\Rend_\theta\left(\{ S[i](\P^k, \D^k,\C^{k,l}) \}_{i=0}^T)\,\right),\;I^{k,l}\right)\,,
    \label{eq:npbg_loss}
    \end{aligned}
\end{equation}

The mismatch $\Delta$ used in NPBG is a perceptual loss evaluating the proximity of the activations of a pretrained VGG network~\cite{Simonyan14}.

\subsection{TRANSPR}

In this subsection, we will describe the differences introduced in our method (TRANSPR), for handling scenes with semi-transparent objects and parts.

\begin{figure*}[t]
  \begin{center}
    \includegraphics[width=1.0\textwidth,clip,trim={0cm 9.5cm 0cm 14cm}]{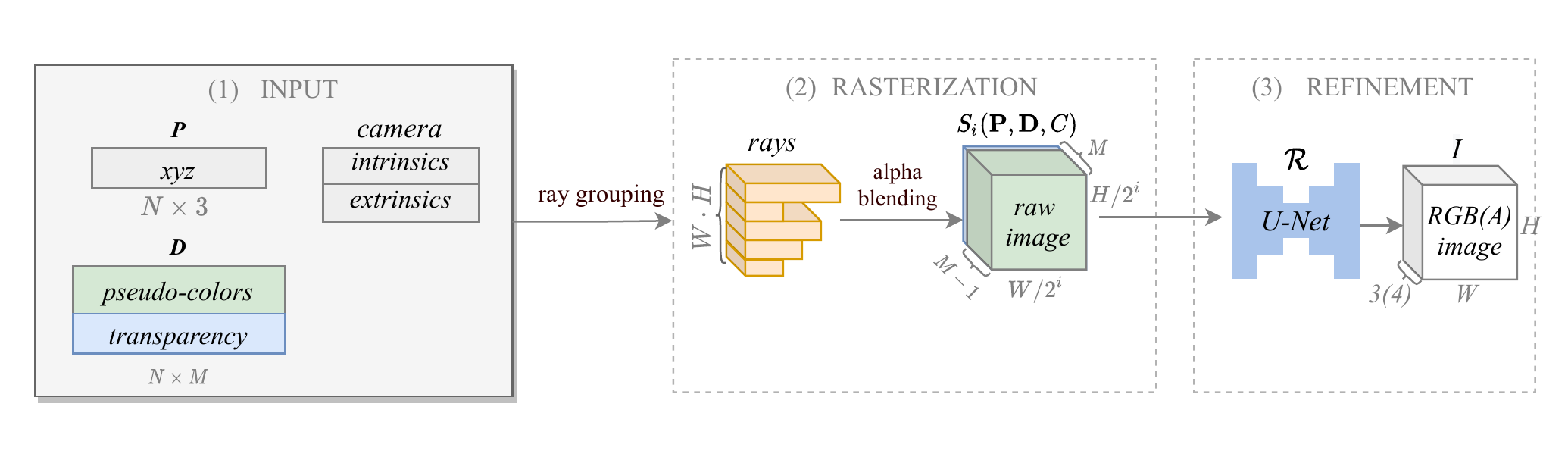}
  \end{center}
  \caption{An overview of the TRANSPR pipeline. At the ray-grouped rasterization step (2), the points are first assigned to a set of $W \cdot H$ rays, with each ray being assigned up to $L$ point descriptors. The result for each ray is aggregated sequentially by blending $(M-1)$-dimensional pseudocolors with the scalar transparency values of its points -- this forms a \textit{raw image}. A pyramid of raw images (not shown in the figure for simplicity) is processed by the rendering network, which can output either RGB image, in case there is no alpha-channel supervision, or RGBA image otherwise. The alpha-channel of the latter can be used for blending with a background image, if needed.}
  \label{fig:scheme_ab}
\end{figure*}

\paragraph{Ray-grouped rasterization.} The rasterization process of NPBG method is the main limitation towards transparent appearance modelling, as only points of the closest to the camera object surfaces get projected during the hard z-buffering. Here we describe an alternative rasterization procedure that accounts for both close and distant points. 



Let us first explain the creation of $S[0](\P, \D, C)$ image of size $W \times H$, and define $W \cdot H$ buckets $r_{ij},\, i \in \{ 1 \dots H\},\, j \in \{1 \dots W\}$ (arrays of potentially variable length), all empty at first. Similarly to the NPBG, we calculate $(x_i, y_i)$ screen-space coordinates for each point $p_i$ on the canvas $S[0]$, as well as its depth $z_i$ w.r.t. the camera $C$, via perspective projection. These screen-space coordinates $(\lfloor x_i \rfloor, \lfloor y_i \rfloor)$ are rounded down, and the descriptor $d_i$ of this point is added to the bucket $r_{\lfloor x_i \rfloor, \lfloor y_i \rfloor}$. The allocation process thus distributes points across buckets of variable length, which we later refer to \textit{rays}. Subsequently, the points of each ray are sorted in order of their increasing depth $z_i$. For all non-empty rays, only the first $L$ descriptors are left and the rest of the points are discarded. The following notation stands for the set of the collected rays:
\begin{align}
    R(\P,\D,\C) = \{r_{11}, r_{12},  \dots, r_{HW} \},
\end{align}
\noindent and let $l_{ij},\, 0 \le l_{ij} \le L$ be a number of points stored in a bucket (ray). Note that some of the rays can remain empty.

\begin{figure}[h]
  \begin{center}
    \includegraphics[width=0.45\textwidth]{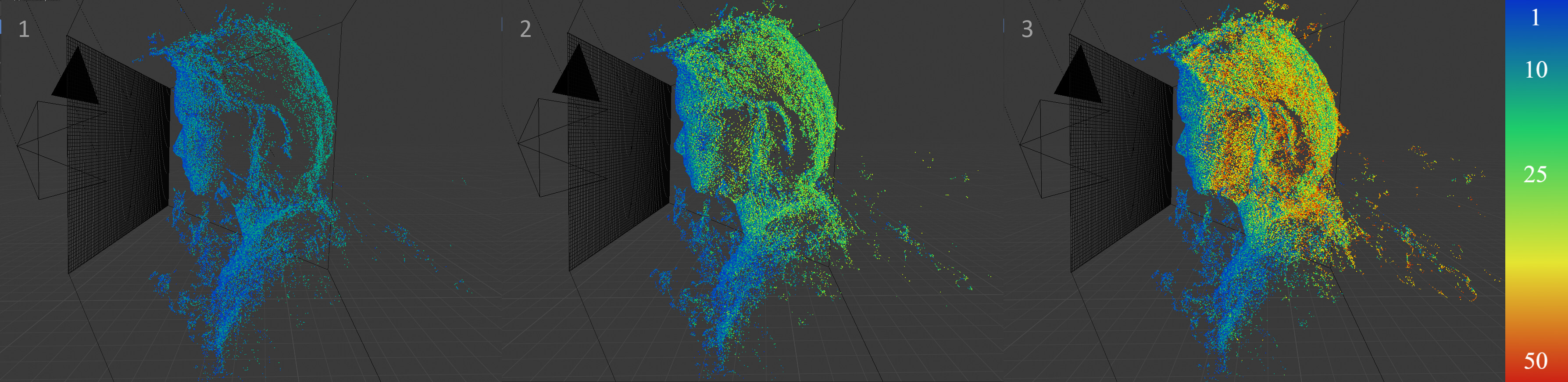}
  \end{center}
  \caption{The illustration of the ray grouping for varying maximum length of ray: (1) $L=10$, (2) $L=20$, (3) $L=50$. Points are either left transparent (if they are discarded with the selected ray length) or colored according to their index on a ray (the mapping is defined on a color bar on the right). Rays are cast from the camera origin (here to the left of the scene); each ray crosses the corresponding pixel of the image canvas.}
  \label{fig:scheme_rays}
\end{figure}

The bucket allocation procedure can be naturally interpreted as the ray grouping. Let us imagine $W \times H$ rays originating at the camera $C$ location and each of them passing through the respective pixel $S[0]_{ij}$ of the image canvas. After the calculation of screen-space coordinates for all points in $\P$, we can attach each point to a ray passing through the screen coordinate of this point. The maximum ray length $L$ is a newly introduced parameter which controls the number of considered surfaces (see Fig.~\ref{fig:scheme_rays}  for an illustration).

\paragraph{Front-to-back alpha blending.} Here we describe the fusion of the multi-slice information stored in $R(\P,\D,\C)$ into a raw image. In TRANSPR, we model not only the non-interpretable descriptors of the points, but also their explicit relative visibility. To support that, we split the $M$-dimensional point descriptor $d_i$ of each point into a concatenation of $(M-1)$-dimensional pseudocolor $c$ (with the same meaning as in NPBG) and a scalar transparency value $\alpha$. At this point, all the transparency values stored in the $R(\P,\D,\C)$ get transformed by the ReLU function and subsequently by tanh, thus constraining $\alpha_i$ to $[0, 1]$. The ReLU+tanh combination is used instead of the sigmoid to allow the outliers to become fully transparent during fitting. The dimensionality $M$ is set to 8, as proposed in the original pipeline \cite{Aliev20}.

The aggregation of $W \cdot H$ rays contained in $R(\P,\D,\C)$ into a raw image $S[0] \in \mathbb{R}^{W \times H \times M}$ is made in a front-to-back manner. Pseudocolors stored in a given ray $r_{ij} = (d_1, \dots, d_{l_{ij}}) = ((c_1, \alpha_1), \dots, (c_{l_{ij}}, \alpha_{l_{ij}}))$ are blended with their transparency values as weights. Aiming for the high level of physical correctness, we sequentially apply the UNDER~\cite{alpha_blend} operator known in computer graphics:

\begin{align}
    \begin{cases}
        C_0 := \boldsymbol{0} \\
        A_0 := 1 \\
        C_k := \alpha_k A_{k - 1} \cdot C_{k-1} + c_k,&  k = 1, \dots, l_{ij} \\
        A_k := (1 - \alpha_k) \cdot A_{k-1},&  k = 1, \dots, l_{ij} \\
        C_{final} := C_{l_{ij}} \\
        A_{final} := 1 - A_{l_{ij}} 
    \end{cases}
    \label{eq:front_to_back_alpha}
\end{align}

The resulting $M$-dimensional vector $D_{l_{ij}} = (C_{final}, A_{final})$ is assigned to $S[0][i, j]$ (consistently with the notation, if the ray was empty, a null descriptor will be assigned). The ray-grouped rasterization is implemented as a fully vectorized sequence of tensor operations and features additional optimizations such as discarding the empty rays, and alpha blending is made slicewise, also via tensor operations on all points simultaneously (to facilitate parallel processing, we pad the rays with fewer than $L$ points with fully transparent points that do not participate in the front-to-back blending). A pyramid of images $\{S[t]\}_{t=1}^T$ is computed precisely the same way with an exception that $\frac W{2^t} \times \frac H{2^t}$ rays are considered.




\paragraph{Rendering network.} The pyramid of raw images $S[0], \dots, S[T]$ is processed by a rendering network~(\ref{eq:rendering_net}) of the same architecture as the one used in NPBG, thus producing the final RGB image $I_{W \times H}$. Fig.~\ref{fig:scheme_ab} depicts the building blocks of TRANSPR.

\subsubsection{Loss functions}

Similarly to NPBG, TRANSPR uses the loss with a VGG-based mismatch $\Delta$ (\ref{eq:npbg_loss}). In case RGB ground truth is only available, the same loss expression as in NPBG is employed. One may consider the case when the rendering network outputs a 4-channel RGBA image $I^{RGBA}_{W \times H}$ instead of a 3-channel RGB $I_{W \times H}$ (by modifying the number of the output channels). Depending on the ground truth, two different loss expressions can be used. When the target image is RGBA too, we penalize RGB and A components separately: 

\begin{equation}
    \begin{aligned}
        \Delta'(I^{RGBA}, \widetilde{I}^{RGBA}) =& \, l_{vgg}(I^{RGB}, \widetilde{I}^{RGB}) \\
        &+ \beta \cdot l_{L1}(I^{A}, \widetilde{I}^{A})
    \label{eq:rgba_loss}
    \end{aligned}
\end{equation}
where $l_{L1}(\cdot, \cdot)$ is the $L_1$ loss over images, $I^{RGB}$ and $I^{A}$ refer to first 3 channels and the $4^{th}$ channel of $I^{RGBA}$, respectively, and $\beta \in \mathbb{R}$ is a predefined parameter. We only make use of the mismatch~(\ref{eq:rgba_loss}) for our experiments with synthetic data, where ground truth alpha can be prerendered.

Another option is to blend the RGBA output with a background image $B^{RGB}$:

\begin{align}
    \hat{I}^{RGB} = (1 - I^{A}) B^{RGB} + I^{RGB}
    \label{methods:bck_blend}
\end{align}

\noindent and compare it with a target RGB image, either captured with background similar to $B^{RGB}$ or obtained from RGBA by the same blending expression. In this scenario, the mismatch $\Delta$ is the same as in NPBG (VGG term only) and is calculated between the blended RGB $\hat{I}^{RGB}$ and the ground truth RGB image. This penalty is going to be useful with real data, when the network is trained to predict the RGBA output, that is blended with background by~(\ref{methods:bck_blend}) and compared with the RGB ground truth, obtained by the hard foreground/blending blending with the same background ($\widetilde{I}^A_{ij} \in \{0, 1\}$).




\subsubsection{Augmentation techniques}
\label{subsubsec:augmentation}

We use several types of train time augmentations. First, the network is trained on random crops (of size 512x512 in our experiments -- can be selected according to the available GPU memory size) sampled from images after random zoom-in (in a range [0.5, 2]x in our experiments), as suggested in ~\cite{Aliev20}. Additionally, we introduce two augmentation strategies which help us perform the scene editing and are not required in other cases. These two techniques are only applied in a setting when the RGBA output is predicted, blended with a background according to the formula~(\ref{methods:bck_blend}), and compared by a VGG loss with the RGB target, blended from RGBA the same way. 

\paragraph{\textit{Alpha jitter.}} This augmentation allows to manipulate transparency of the objects after training by rescaling the alpha values of these objects. For each training batch, we choose random number of samples that will be altered. Let us say that $p$ is a value between 0 and 1, uniformly sampled for each altered training sample in the batch, and a newly introduced scalar scene-wise parameter $\mu$ is gradually optimized during the training process (if several scenes are fitted, we optimize single $\mu_k$ per scene). Before the front-to-back blending~(\ref{eq:front_to_back_alpha}), we rescale the alpha values contained in each ray $r_{ij} = ((c_1, \alpha_1), \dots, (c_{l_{ij}}, \alpha_{l_{ij}}))$. The following rescaling of alpha values is applied to each of these rays:

\begin{align}
    \hat{\alpha}_k = \alpha_k \cdot p^\mu,\,\, k = 1, \dots, l_{ij}
    \label{eq:jitter}
\end{align}

The values $\alpha_k$ in a ray are replaced by the rescaled ones ($\hat{\alpha}_k$). Accordingly, the target alpha channel used to produce the target image is multiplied by $p$ ($\widetilde{I}^A_{ij} \in \{0, p\}$). This procedure prepares the network for subsequent transparency manipulation. At the inference time, $p$ can be selected and applied using the expression~(\ref{eq:jitter}) for those points, transparency of which needs to altered. The exponent $\mu$ enables the higher visual quality in our experimental setting, since when $\mu > 1$, it corresponds to the slower accumulation of the blended alpha $A_k$~(\ref{eq:front_to_back_alpha}), which makes the opacity effect more apparent when the number of points in a ray is large.   



\paragraph{\textit{Overlay.}}

In order to learn the transparency concept for scene editing from a limited number of scenes, we augment training data by overlaying random pairs of samples in batch. 

The augmentation is applied when the network is trained on two scenes that will later be rendered together. Suppose that a random training batch contains samples corresponding to the two views $C_{f_{j}}, C_{b_{k}}$ of either the same scene or distinct scenes. As the scenes and views are sampled randomly, we artificially assume (for the sake of augmentation) that the perceived object at the first view is completely frontal compared to the perceived object at the second view. First, raw images $S[i]_{f_{j}} = [C[i]_{f_{j}}, A[i]_{f_{j}}],\, S[i]_{b_{k}} = [C[i]_{b_{k}}, A[i]_{b_{k}}]$ are produced independently for each view. Later, the described assumption is applied by blending them w.r.t. their alpha channels:

\begin{align}
    \hat{C}[i] = C[i]_{f_{j}} +  C[i]_{b_{k}} \cdot (1-A[i]_{f_{j}}) \cdot  A[i]_{b_{k}} \\
    \hat{A}[i] = 1-(1-A[i]_{b_{k}})\cdot (1-A[i]_{f_{j}}),
    \label{eq:overlay}
\end{align}

\noindent and modified raw images $\{\hat{S}[i] = [\hat{C}[i], \hat{A}[i]]\}_i$ are passed to the rendering network. Target images are transformed similarly:

\begin{align}
    \hat{I}^{RGB} =  I^{RGB}_{f_{j}} +   I^{RGB}_{b_{k}} \cdot (1 - I^{A}_{f_{j}}) \cdot  I^{A}_{b_{k}} \\
    \hat{I}^{A} = 1-(1-I^{A}_{b_{k}})\cdot (1-I^{A}_{f_{j}})
    \label{eq:overlay}
\end{align}

Since the same alpha-blending rule is applied as for the rays~(\ref{eq:front_to_back_alpha}), the strategy corresponds to the front vs. back UNDER-operator based overlay. 


\section{Experiments}


\subsection{Datasets}
In our experiments, we use synthetic and real data to demonstrate both the pipeline representation power for semi-transparent scenes and its suitability for real scenes.

\subsubsection{Synthetic}\label{data:synth}

Synthetic datasets were rendered using Blender Eevee rendering engine. Multi-view imagery was captured by sampling cameras on a sphere/hemisphere around a scene. Cameras share the same intrinsic parameters, while each image has a corresponding extrinsic (view) matrix. The following scenes were considered:

\begin{figure}[h]
  \begin{center}
    \includegraphics[width=0.48\textwidth]{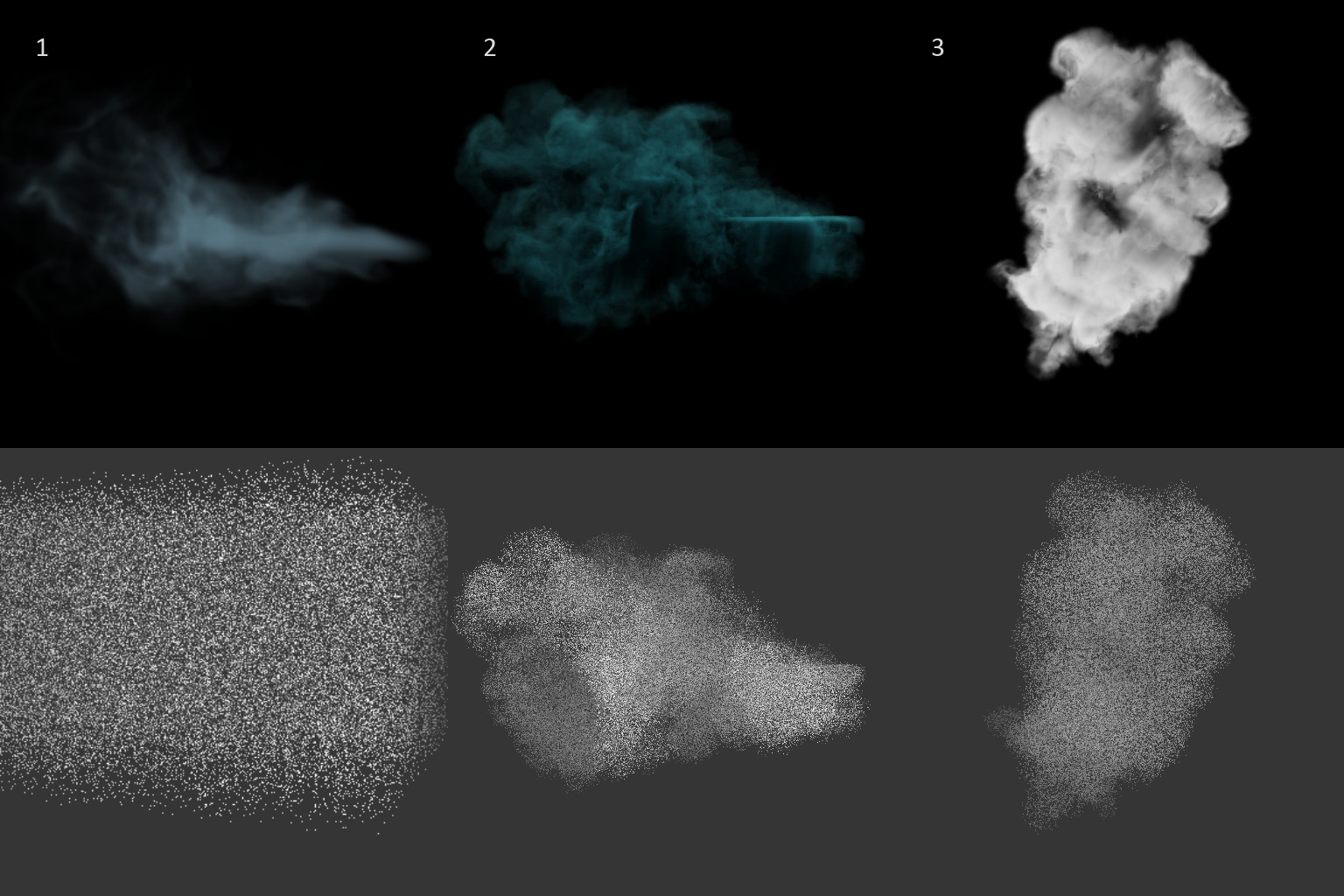}
  \end{center}
  \caption{Captures from the synthetic smoke datasets. Left to right: (1) Dynamic smoke, (2) Static smoke, (3)  a sample from Clouds collection.}
\end{figure}

\textit{Aquarium}. A semi-transparent scene of the aquarium with fish\footnote{model available at \url{free3d.com}, distributed with Royalty Free License allowing (non-)commercial use} (without any refraction/reflection effects) with 100 RGBA images from different viewpoints. The point cloud of $10^6$ points has been sampled from the corresponding meshes.  

\textit{Dynamic smoke}. A dynamic over time, 50-timeframe long animation of monochrome smoke without lighting or shadows, rendered in RGBA from 64 static cameras. Each frame corresponds to a different temporal state of the simulation. A rectangular voxel volume served as a geometric proxy for the TRANSPR where 400'000 points are sampled within the volume.\footnote{the technique is outlined in the Blender Made Easy tutorial:\linebreak \url{https://www.youtube.com/watch?v=_IDBfQnttiE}} 

\textit{Static smoke / Clouds}. This collection of datasets is comprised of several static (not time-dependent) scenes of a smoke/cloud of different shapes with volumetric lighting and shadows turned on. Each scene is based on 80 multi-view RGBA images. Point clouds were obtained by thresholding and sparsifying voxel density field for each of the smoke simulation objects.

\textit{Aquarium} and \textit{Dynamic smoke} synthetic scenes were rendered at medium pixel resolution of $512 \times 512$. \textit{Static smoke} and \textit{Clouds} scenes were used in composition with real data and were rendered at $1920 \times 1920$. Holdout sequences of 400 frames in total were rendered for \textit{Aquarium} and \textit{Dynamic smoke} (for the latter, images were taken at every second timestamp; the others were selected as training). These novel views were unseen during training and used for quantitative and visual evaluation (see \ref{exp:synth}). 

\subsubsection{Real}\label{data:real}


All datasets except the drone footage (\textit{Church}) were collected by either mobile or DSLR cameras and contain semi-transparent materials, such as chiffon or glass. The images were captured using $180^\circ$ or $360^\circ$ smooth trajectories resulting into $\sim$130 frames with a small deviation for each of the datasets. \textit{Chiffon shirt}, \textit{Flowers in a vase}, \textit{Jewelry,} and \textit{Aloe} \cite{Aliev20} scenes were captured using a mobile camera, and their photos were subsequently downsampled to $1080 \times 1920$ resolution. \textit{Scarf} and \textit{Table} scenes were captured by a DSLR camera and downsampled to $1080 \times 1620$ resolution. 
\textit{Church} scene was assembled from the DJI Phantom 4 Pro flying drone footage (126 frames downsampled to $2736 \times 1824$ pixels) to demonstrate the pipeline capability to augment real landscapes with atmospheric visual effects. 

The point clouds were obtained by a photogrammetric software package Agisoft Metashape \cite{agisoft}. Since transparent glass is an extremely challenging surface for such reconstruction, a special scenario with two sequences was considered for the \textit{Flowers in a vase} scene. First, a $180^{\circ}$ sequence of flowers in a transparent vase was captured as is. Afterwards, the vase was wrapped into a paper with checkerboard pattern. Finally, point clouds of each sequence were reconstructed (one with the vase not reconstructed, and another featuring the vase in the scene) and geometrically aligned. Since the RGB point colors are not used by both TRANSPR and NPBG, they can be safely discarded. To model the appearance of this scene, the TRANSPR is given a combined point cloud as an input and photographs with the registered camera poses of the first sequence as the ground truth.


\subsection{Synthetic objects. RGBA supervision}\label{exp:synth}
The main motivation for the experiments on \textit{Aquarium} and \textit{Dynamic smoke} scenes was to evaluate the pipeline capability of learning physically accurate transparency in a fully supervised mode, i.e., when both target and model output are 4-channel RGBA.

In addition to the comparison with NPBG \cite{Aliev20}, we compare the performance of TRANSPR against current state-of-the-art neural volumetric rendering methods for novel view synthesis \cite{Lombardi19,Mildenhall20}. These approaches are image-based, i.e.,\ only require registered multi-view images as inputs for the fitting of each scene (as opposed to NPBG and TRANSPR that also require a point cloud).

We report \textit{L1}, \textit{VGG loss}~\cite{Johnson16} and \textit{LPIPS}~\cite{Zhang18} for the quantitative evaluation of the results.

\paragraph{Setup.} To ensure a fair comparison to NPBG on the \textit{Aquarium} scene, the rendering network in the NPBG pipeline was slightly modified to output a 4-channel RGBA image instead of a 3-channel RGB. In conjunction with that, the loss with a separate alpha term~(\ref{eq:rgba_loss}) was used.

For the \textit{Dynamic smoke} experiment, every second animation frame was used for training to investigate the ability of methods to recover intermediate time-frames. A temporal state for TRANSPR is associated with a set of neural descriptors values trained on imagery for a corresponding time frame. The use of the point cloud with the same point locations for all time frames allowed to interpolate point descriptors to produce intermediate states. Since we investigate the generation of every second frame, linear interpolation in the descriptor space was used in our experimental setting, yet more complex interpolation strategies are possible (e.g.~cubic interpolation). This is the only experiment where we used raw alpha values without normalizing transformations to prevent flickering artifacts on interpolation. \textit{Neural Volumes}~\cite{Lombardi19} uses a view-conditioning strategy that generates scene volume for a particular temporal state based on a subset of 3 train cameras. \textit{NeRF} does not allow to directly interpolate neural representation, since a separate neural network is trained for each time frame. The inferred images for the trained frames were linearly interpolated instead.

The maximum ray length for TRANSPR was set to 50 for both scenes.

\begin{figure}[h]
  \begin{center}
    \includegraphics[width=0.49\textwidth]{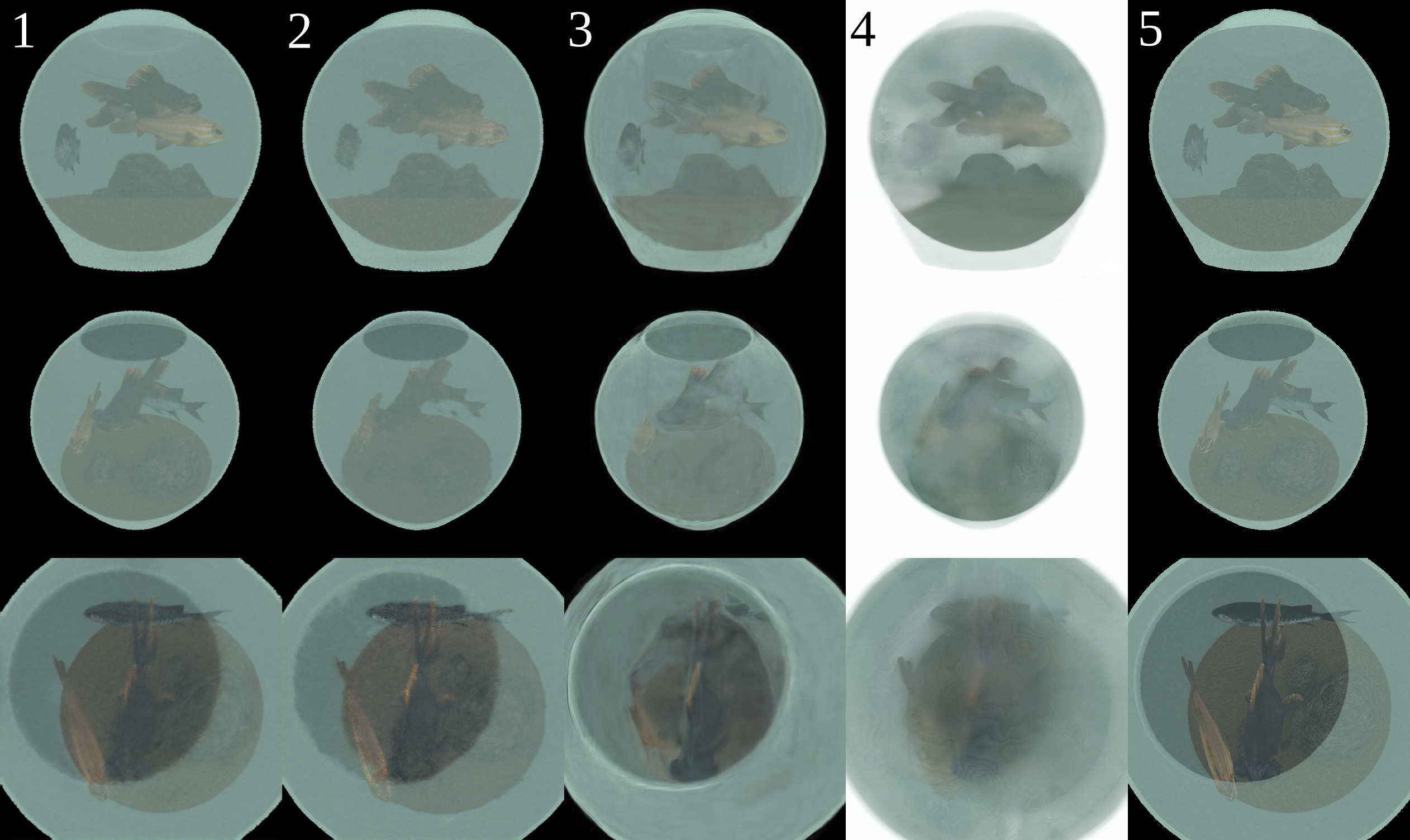}
  \end{center}
  \caption{The comparative performance on the \textit{Aquarium} scene. Novel views rendering. Left to right: (1) TRANSPR (Ours), (2) NPBG, (3) Neural Volumes (4) NeRF, (5) ground truth. \textit{Electronic zoom-in recommended.}}
  \label{fig:res_aquarium}
\end{figure}

\paragraph{Results.} Fig.~\ref{fig:res_aquarium} contains the visual quality assessment for the \textit{Aquarium} scene. In addition, we report the qualitative comparison of all methods for \textit{Aquarium} and \textit{Dynamic Smoke} in Table~\ref{tables:synth_metrics}. TRANSPR  outperformed image-based approaches, as well as the NPBG baseline on the \textit{Aquarium} scene. While NPBG lacks contextual information in the raw images due to the visible scene truncation as a result of hard z-buffering, TRANSPR efficiently handles multi-surface point cloud structure. Although TRANSPR lacks fine details on the \textit{Dynamic smoke} scene, it demonstrates robustness to the change of viewpoints and proves the ability to synthesize intermediate temporal states of the simulation. In addition, the alpha channel predicted by TRANSPR enables straightforward background blending (see Fig.~\ref{methods:bck_blend}).

NeRF demonstrates poor generalization to non-train views on our synthetic scene, which results in blurry holdout renderings for all scenes. Both Neural Volumes and NeRF produce specific artifacts on the \textit{Aquarium} scene. Nevertheless, Neural Volumes proves better detailization on \textit{Dynamic smoke}. Still, this method requires a set of training images preloaded at the inference time for view conditioning, which can be impractical in cases when memory is limited. Indeed, for a relatively small point cloud of dynamic smoke consisting of 0.4 million points, TRANSPR requires 30 MB of GPU memory at inference time, while Neural Volumes takes 630 MB. 

Additionally, TRANSPR is faster compared to the image-based methods, as our implementation yields four FPS on the NVIDIA Tesla P100 GPU, while the trained model by Neural Volumes can be executed only at 1 FPS and the NeRF model runs at 0.08 FPS on the same hardware.

\begin{figure}[h]
  \begin{center}
    \includegraphics[width=0.49\textwidth,clip,trim={0 10cm 0 0}]{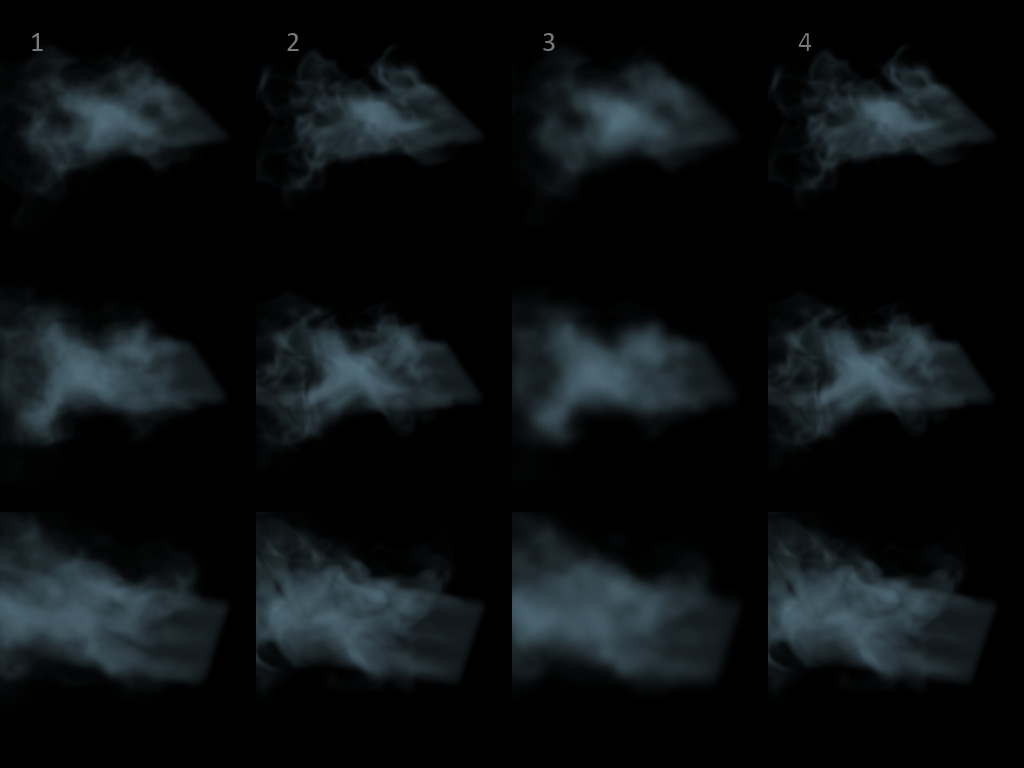}
  \end{center}
  \caption{The comparative performance on the \textit{Dynamic smoke} scene. For each of the three rows, novel view is generated for some intermediate holdout timestamp of the dynamic animation, while the previous and next timestamp were used for training. Left to right: (1) TRANSPR, (2) Neural Volumes, (3) NeRF, (4) ground truth. \textit{Electronic zoom-in recommended.}} 
\end{figure}


\begin{table}[h]
\centering 
\resizebox{0.47\textwidth}{!}{
\begin{tabular}{r|c|c|c|c|c|c}
                                 & \multicolumn{3}{c|}{\textit{Aquarium}}                           & \multicolumn{3}{c}{\textit{Dynamic Smoke}}                                      \\ \hline
Method                           & VGG $\downarrow$ & L1 $\downarrow$ & LPIPS $\downarrow$ & VGG $\downarrow$       & L1 $\downarrow$        & LPIPS $\downarrow$    \\ \hline
TRANSPR (Ours)                 & \textbf{120.89}  & \textbf{.0112} & \textbf{.1061}    & \textbf{117.35}        & .0053                 & .0432                \\ \hline
NPBG (RGBA) & 143.22           & .0134          & .1238             & --- & --- & --- \\ \hline
Neural Volumes                 & 209.01           & .0172          & .2202             & 118.05               & .0051                 & \textbf{.0429}       \\ \hline
NeRF                             & 327.60           & .0551          & .3220             & 117.53                 & \textbf{.0047}        & .0669                \\ \hline
\end{tabular}
}
\caption{Quantitative results for the compared methods on \textit{Aquarium} and \textit{Dynamic smoke} scenes. "---" indicates that the method is not applicable to the scene (in particular, NPBG is unable to render a semi-transparent smoke represented as a dense grid of points in a volume).}
\label{tables:synth_metrics}
\end{table}

\subsection{Real-world objects. RGB supervision}\label{exp:real_rgb}

In this subsection, we show the capabilities of TRANSPR at realistic modeling of scenes with semi-transparent parts, combining them, augmenting with synthetic data, and manipulating their appearance by changing learned pointwise opacity on the data captured in in-the-wild conditions.

\subsubsection{Modeling semi-transparency}

In contrast to synthetic scenes, real scenes do not come with the ground truth transparency, which makes the learning process more tricky.

\begin{figure*}[t!]
    \begin{minipage}{\textwidth}
        \centering
        \includegraphics[width=.9\textwidth]{\detokenize{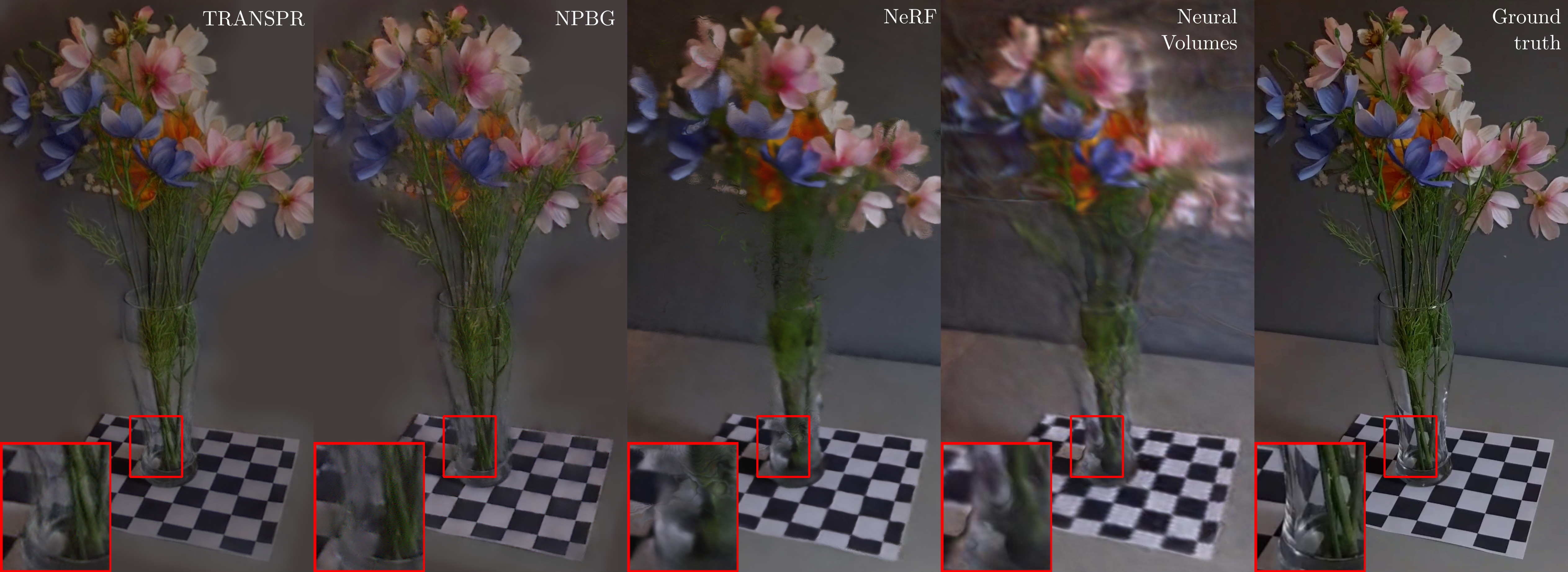}}
        \caption{Comparison of the \textit{Flowers in a vase} transparent scene rendering. Each picture in a bottom-left corner is a zoom-in of a region of interest which belongs to the transparent vase and the flower stems. Left-to-right: (1) TRANSPR, (2) NPBG, (3) NeRF, (4) Neural Volumes, (5) ground truth. \textit{Electronic zoom-in recommended.}}
        \label{fig:real_flowers}
    \end{minipage}
    \vspace{0.1cm}
    \begin{minipage}{\textwidth}
        \begin{center}
            \includegraphics[width=.85\linewidth]{\detokenize{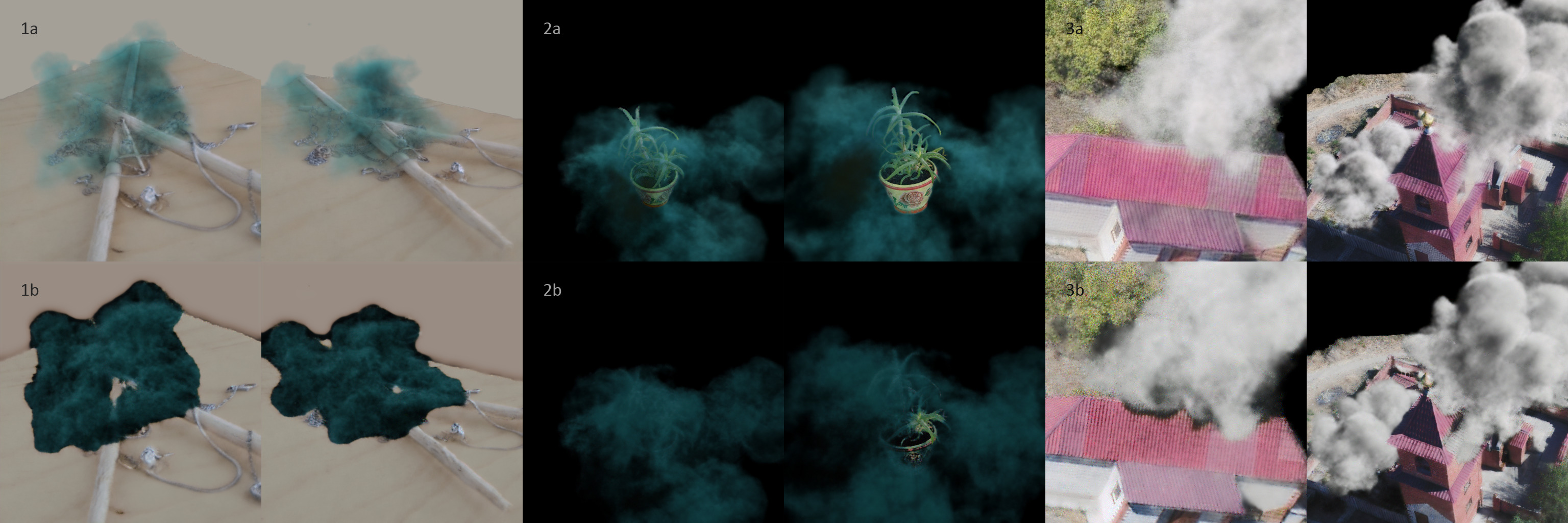}}
          \end{center}
          \caption{Scene composition. Top row: TRANSPR (a), bottom row: NPBG (b). (1) \textit{Static smoke} and \textit{Jewelry}, (2) \textit{Static smoke} and \textit{Aloe}, (3) \textit{Clouds} collection and \textit{Church}. \textit{Electronic zoom-in recommended.}}
          \label{fig:edt_with_smoke}
    \end{minipage}
    \vspace{0.1cm}
    \begin{minipage}{\textwidth}
        \begin{center}
            \includegraphics[width=.8\textwidth]{\detokenize{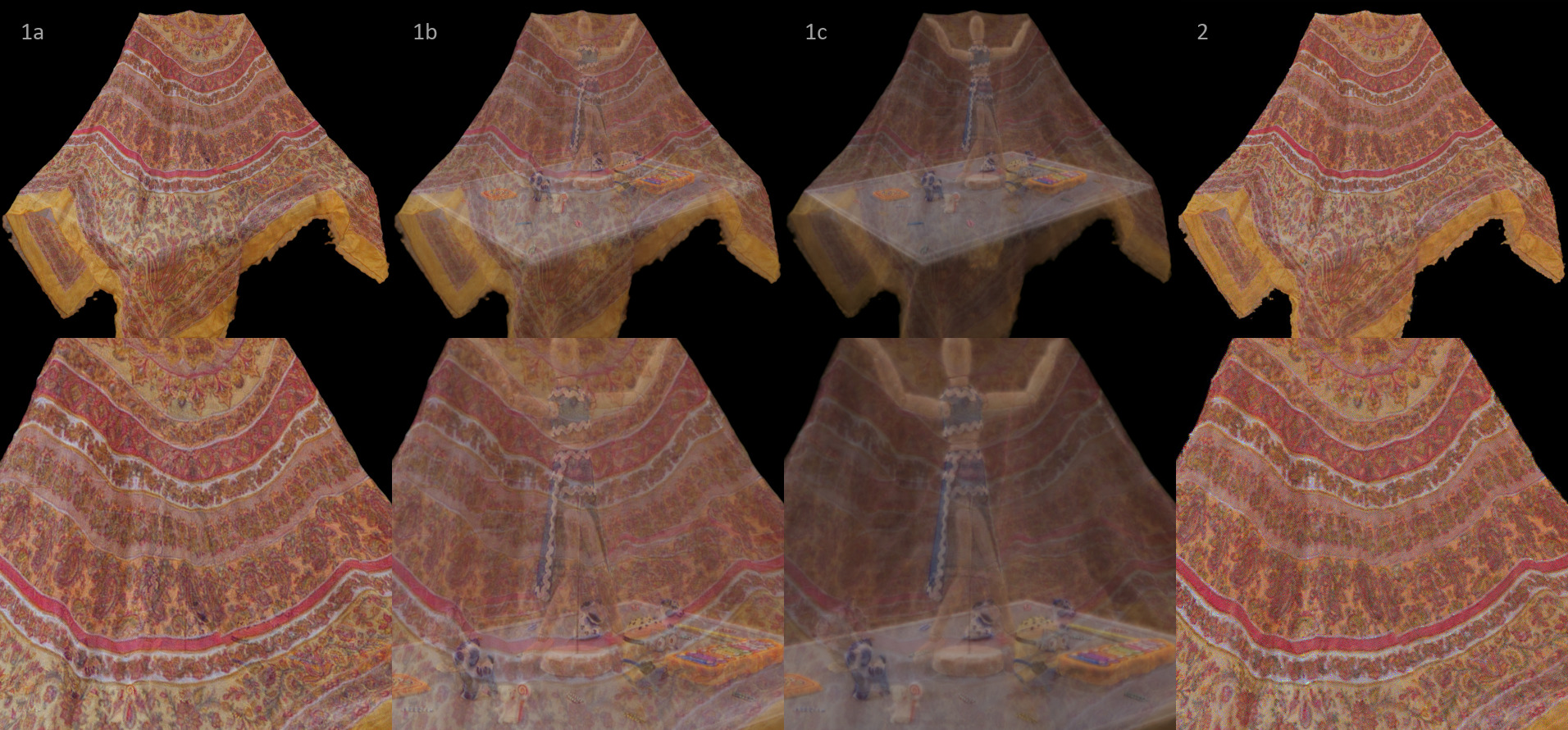}}
        \end{center}
        \caption{Scene editing with \textit{Scarf} and \textit{Table} scenes. TRANSPR with (1a) alpha as learned, (1b) alpha decreased by a multiplying factor of 0.6, (1c) alpha decreased by a multiplying factor of 0.35; (2) NPBG. }
        \label{fig:edt_scarf_table}
    \end{minipage}
\end{figure*}

The renderings of \textit{Flowers in a vase} are depicted in Fig.~\ref{fig:real_flowers} and supported by the metrics in Table~\ref{tables:flowers_metrics}. All metrics were calculated for the segmented predicted and ground truth images; the segmentation masks were prepared manually using Adobe Photoshop CC software package (the masks were not used during fitting). Compared to NPBG, TRANSPR ($L=30$) efficiently handles nested structure of the point cloud and demonstrates robustness to novel views. NPBG is still able to produce plausible results, especially for viewpoints close to the model, as it is able to use surface bleeding to its advantage: points descriptors of the farther surfaces are getting projected between the closer points when the object occupies a large space on the canvas. Nevertheless, NPBG results exhibit more artefacts compared to TRANSPR, including color noise and temporal flickering,  which are absent or less pronounced in the case of TRANSPR (see Supplemental Video for further comparison). NeRF was trained for 500'000 steps and Neural volumes for 46'000 at the original resolution. Both methods required scaling extrinsics by 24 and 32 respectively (scaling factors were taken small enough, such that the whole scene space is covered). In general, it produces more blurry pictures featuring much less detail, which is observed both visually and supported by the higher result of L1 score that is well-known to encourage blurry pictures (see, e.g.,~\cite{Zhang18}). The same applies to Neural Volumes, renderings of which also possess specific artefacts similar to motion blur.  

\begin{table}[h]
    \centering 
    \begin{tabular}{l|c|c|c}
        \multicolumn{1}{r|}{}       & \multicolumn{3}{c}{\textit{Flowers in a vase}}         \\ \hline
        \multicolumn{1}{r|}{Method} & VGG $\downarrow$ & L1 $\downarrow$ & LPIPS $\downarrow$ \\ \hline
        TRANSPR (Ours)              & \textbf{135.42}    &  .0152         &  \textbf{.0625}            \\ \hline
        NPBG                        & 153.61    &  .0176         &  .0702            \\ \hline
        NeRF                        & 142.56    &  \textbf{.0072}         &  .0654  \\ \hline
        Neural Volumes              & 210.24    &  .0146         &  .1163  \\ \hline
    \end{tabular}
    \caption{Quantitative results for the compared methods on \textit{Flowers in a vase} scene.}
    \label{tables:flowers_metrics}
\end{table}


\subsubsection{Transparency manipulation}

We demonstrate the possibility to alter the learned transparency of the objects on \textit{Chiffon shirt} and \textit{Scarf} scenes. Each of them was trained with background blending that required binary silhouettes of the point clouds and a set of random background images applied during training (100 stock images were used in our case, yet any number of arbitrary images could be taken instead). These two scenes do not include any nested geometry, and the maximum ray length was therefore set to a low value of 25 accordingly. Both jitter and overlay augmentations were employed. The way the manipulation is applied is described in the Method part, Subsubsection~\ref{subsubsec:augmentation}.

\begin{figure}[h]
  \begin{center}
    \includegraphics[width=0.45\textwidth]{\detokenize{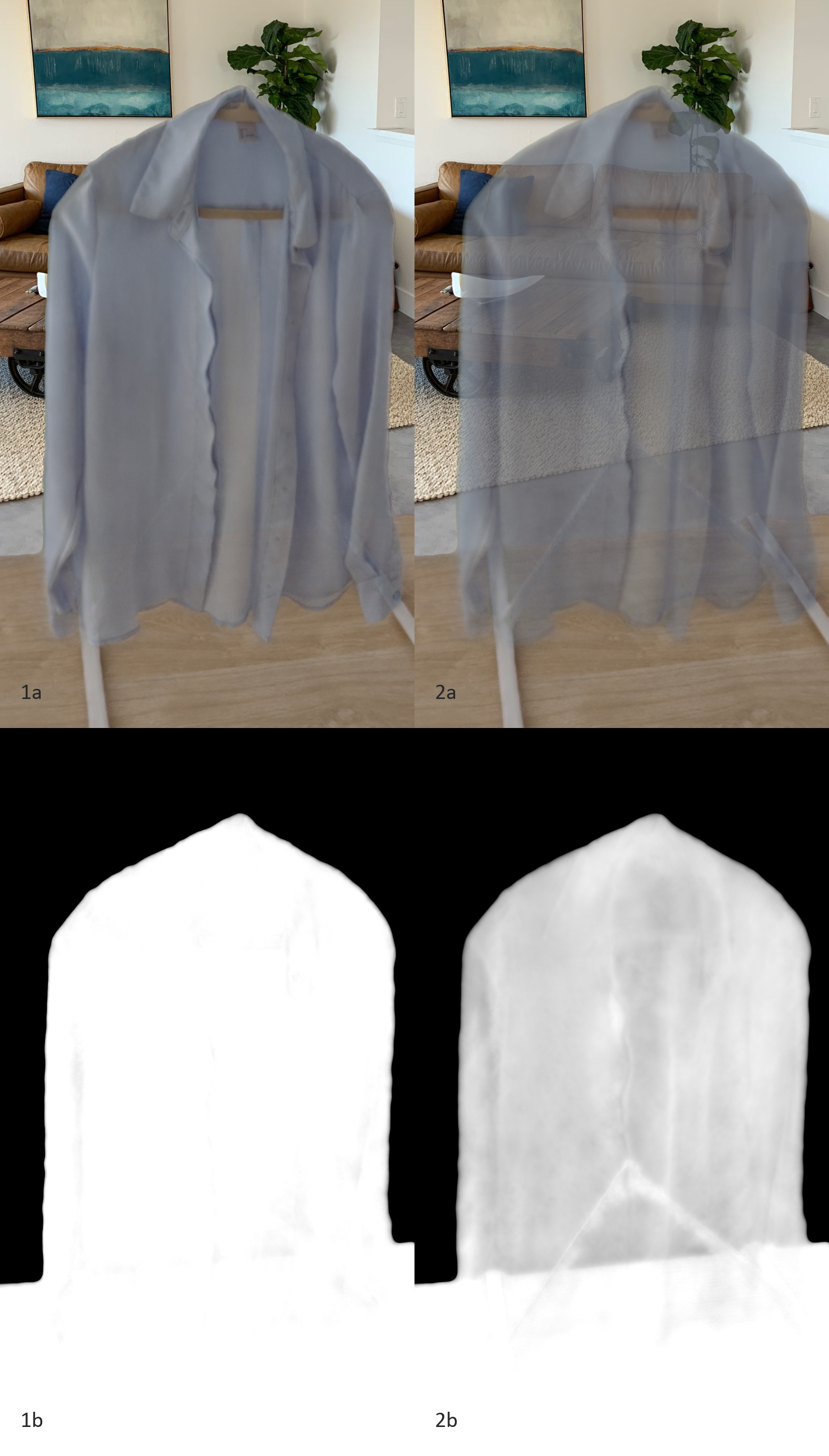}}
  \end{center}
  \caption{The results of transparency manipulation on the \textit{Chiffon shirt} scene. Left column: rendering with alpha as learned. Right column: alpha values of \textit{Chiffon shirt} decreased by a factor of 0.65. Top row: \textit{RGB} channels of the rendered image, bottom row: \textit{A} channel of the rendered image.}
  \label{fig:manip_shirt}
\end{figure}

\begin{figure}[h]
  \begin{center}
    \includegraphics[width=0.45\textwidth]{\detokenize{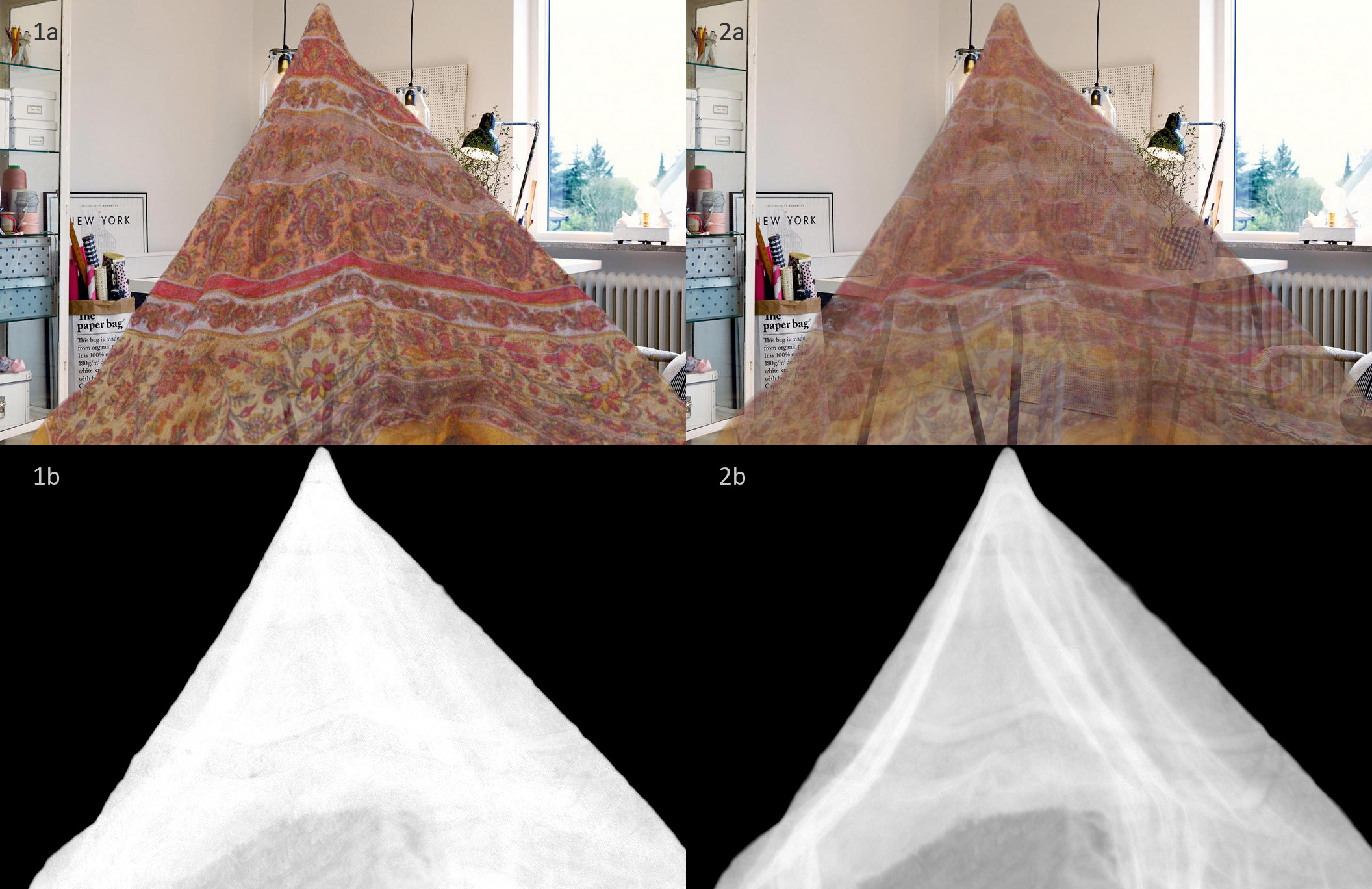}}
  \end{center}
  \caption{The results of transparency manipulation on the \textit{Scarf} scene. Left column: rendering with alpha as learned. Right column: alpha values of \textit{Scarf} decreased by a factor of 0.6. Top row: \textit{RGB} channels of the rendered image, bottom row: \textit{A} channel of the rendered image.}
  \label{fig:manip_scarf}
\end{figure}

The results in Fig.~\ref{fig:manip_shirt} for \textit{Scarf} and in Fig.~\ref{fig:manip_scarf} \textit{Chiffon Shirt} demonstrate both photorealistic transparency blending, especially pronounced in the case of \textit{Chiffon Shirt}. Note that texture subtleties are preserved during the manipulation.

\subsubsection{Scene composition}

In addition, we show how TRANSPR extends the scene editing scenario originally proposed as a use case for NPBG in~\cite{Aliev20}. In the case of TRANSPR, one can compose the objects with added or altered transparency, and jointly render synthetic assets with the data from the real world. 

We train a separate model for a pair of scenes using RGB supervision for both methods. Then, we geometrically align the point clouds and render them such that each point is equipped with a descriptor learned after the corresponding fitting stage (either the first or the second). The maximum ray length for TRANSPR is set to 50. Fig.~\ref{fig:edt_scarf_table} showcases a composition of \textit{Scarf} and \textit{Table} with both $\alpha$ as learned and with decreased for the exterior (\textit{Scarf}), as well as the comparison with NPBG which can only render both objects at a closer look. The composition of \textit{Scarf}+\textit{Table} was trained with both jitter and overlay augmentations. Fig.~\ref{fig:edt_with_smoke} and Fig.~\ref{fig:teaser} contain scene compositions with semi-transparent \textit{Static smoke} covering other objects (see further comparison in the Supplemental Video). These combined renderings were prepared with overlay augmentation enabled during training.

\section{Discussion}

We have presented a neural pipeline for appearance modeling of semi-transparent scenes represented as point clouds. Leveraging ideas from the existing literature, we follow the approach of associating every point with a learnable descriptor and strengthen it by introducing the point-level transparency value. We demonstrate the ability of our method to synthesize photorealistic novel views for scenes with semi-transparent parts, including scenes comprising synthetic and real-world objects. In addition, we show that the learned opacity information, associated with every point, can be jointly altered to blend some parts of the scene with others and to achieve the desired visual effects. 

The limitations of TRANSPR include the inability to model refraction. Furthermore, unlike neural methods based on volume rendering and similarly to NPBG, TRANSPR relies on the externally-provided point cloud. At the same time, such explicit specification of geometry allows TRANSPR to run at higher framerates.



\FloatBarrier

{\small
\bibliographystyle{ieee}
\bibliography{refs}

\begin{thebibliography}{10}\itemsep=-1pt

\bibitem{agisoft}
{Agisoft}.
\newblock {\em Metashape software}, retrieved 20.05.2019.

\bibitem{Aliev20}
K.-A. Aliev, A.~Sevastopolsky, M.~Kolos, D.~Ulyanov, and V.~Lempitsky.
\newblock Neural point-based graphics.
\newblock In {\em Proc. {ECCV}}, 2020.

\bibitem{Bui18}
G.~Bui, T.~Le, B.~Morago, and Y.~Duan.
\newblock Point-based rendering enhancement via deep learning.
\newblock {\em The Visual Computer}, 34(6-8):829--841, 2018.

\bibitem{alpha_blend}
A.~Dunn.
\newblock Transparency (or translucency) rendering,
  https://developer.nvidia.com/content/transparency-or-translucency-rendering,
  2014.

\bibitem{Flynn19}
J.~Flynn, M.~Broxton, P.~Debevec, M.~DuVall, G.~Fyffe, R.~Overbeck, N.~Snavely,
  and R.~Tucker.
\newblock Deepview: View synthesis with learned gradient descent.
\newblock In {\em Proc. {CVPR}}, pages 2367--2376, 2019.

\bibitem{Gross02}
M.~Gross, H.~Pfister, M.~Alexa, M.~Pauly, M.~Stamminger, and M.~Zwicker.
\newblock {\em Point based computer graphics}.
\newblock Eurographics Assoc., 2002.

\bibitem{Johnson16}
J.~Johnson, A.~Alahi, and L.~Fei{-}Fei.
\newblock Perceptual losses for real-time style transfer and super-resolution.
\newblock In {\em Proc. {ECCV}}, pages 694--711, 2016.

\bibitem{Kobbelt04}
L.~Kobbelt and M.~Botsch.
\newblock A survey of point-based techniques in computer graphics.
\newblock {\em Computers \& Graphics}, 28(6):801--814, 2004.

\bibitem{Lombardi19}
S.~Lombardi, T.~Simon, J.~Saragih, G.~Schwartz, A.~Lehrmann, and Y.~Sheikh.
\newblock Neural volumes: Learning dynamic renderable volumes from images.
\newblock {\em ACM Transactions on Graphics (TOG)}, 38(4):65, 2019.

\bibitem{Meshry19}
M.~Meshry, D.~B. Goldman, S.~Khamis, H.~Hoppe, R.~Pandey, N.~Snavely, and
  R.~Martin-Brualla.
\newblock Neural rerendering in the wild.
\newblock In {\em Proc. {CVPR}}, June 2019.

\bibitem{Mildenhall20}
B.~Mildenhall, P.~P. Srinivasan, M.~Tancik, J.~T. Barron, R.~Ramamoorthi, and
  R.~Ng.
\newblock Nerf: Representing scenes as neural radiance fields for view
  synthesis.
\newblock In {\em Proc. {ECCV}}, 2020.

\bibitem{Pittaluga19}
F.~Pittaluga, S.~J. Koppal, S.~B. Kang, and S.~N. Sinha.
\newblock Revealing scenes by inverting structure from motion reconstructions.
\newblock In {\em Proc. {CVPR}}, June 2019.

\bibitem{Ronneberger15}
O.~Ronneberger, P.~Fischer, and T.~Brox.
\newblock U-net: Convolutional networks for biomedical image segmentation.
\newblock In {\em International Conference on Medical image computing and
  computer-assisted intervention}, pages 234--241. Springer, 2015.

\bibitem{Simonyan14}
K.~Simonyan and A.~Zisserman.
\newblock Very deep convolutional networks for large-scale image recognition.
\newblock {\em CoRR}, abs/1409.1556, 2014.

\bibitem{Sitzmann19a}
V.~Sitzmann, J.~Thies, F.~Heide, M.~Nie{\ss}ner, G.~Wetzstein, and
  M.~Zollh{\"{o}}fer.
\newblock Deepvoxels: Learning persistent 3d feature embeddings.
\newblock In {\em Proc. {CVPR}}, 2019.

\bibitem{Sitzmann19b}
V.~Sitzmann, M.~Zollh{\"o}fer, and G.~Wetzstein.
\newblock Scene representation networks: Continuous 3d-structure-aware neural
  scene representations.
\newblock pages 1121--1132, 2019.

\bibitem{Szeliski98}
R.~Szeliski and P.~Golland.
\newblock Stereo matching with transparency and matting.
\newblock In {\em Proc. {ICCV}}, pages 517--524. IEEE, 1998.

\bibitem{Tewari20}
A.~Tewari, O.~Fried, J.~Thies, V.~Sitzmann, S.~Lombardi, K.~Sunkavalli,
  R.~Martin-Brualla, T.~Simon, J.~Saragih, M.~Nießner, R.~Pandey, S.~Fanello,
  G.~Wetzstein, J.-Y. Zhu, C.~Theobalt, M.~Agrawala, E.~Shechtman, D.~B.
  Goldman, and M.~Zollhöfer.
\newblock State of the art on neural rendering, 2020.

\bibitem{Thies19}
J.~Thies, M.~Zollh{\"{o}}fer, and M.~Nie{\ss}ner.
\newblock Deferred neural rendering: Image synthesis using neural textures.
\newblock In {\em Proc. {SIGGRAPH}}, 2019.

\bibitem{Yifan19}
W.~Yifan, F.~Serena, S.~Wu, C.~Öztireli, and O.~Sorkine-Hornung.
\newblock Differentiable surface splatting for point-based geometry processing.
\newblock {\em ACM Transactions on Graphics}, 38(6):1–14, Nov 2019.

\bibitem{Zhang18}
R.~Zhang, P.~Isola, A.~A. Efros, E.~Shechtman, and O.~Wang.
\newblock The unreasonable effectiveness of deep features as a perceptual
  metric.
\newblock In {\em Proc. {CVPR}}, 2018.

\bibitem{Zhou18}
T.~Zhou, R.~Tucker, J.~Flynn, G.~Fyffe, and N.~Snavely.
\newblock Stereo magnification: learning view synthesis using multiplane
  images.
\newblock {\em ACM Transactions on Graphics (TOG)}, 37(4):1--12, 2018.

\end{thebibliography}
}
\end{document}